\documentclass[final]{cvpr}

\usepackage{times}
\usepackage{epsfig}
\usepackage{graphicx}
\usepackage{amsmath}
\usepackage{amssymb}

\usepackage{booktabs}
\usepackage{multirow}
\usepackage{enumitem}

\usepackage[pagebackref=false,breaklinks=true,colorlinks,bookmarks=false]{hyperref}

\def\cvprPaperID{313}
\def\confYear{CVPR 2021}

\newcommand{\figLabel}{Figure\xspace}
\newcommand{\eqLabel}{Equation\xspace}
\newcommand{\secLabel}{Section\xspace}
\newcommand{\tblLabel}{Table\xspace}
\newcommand{\mysection}[1]{\vspace{3pt}\noindent\textbf{#1.}}

\newcommand\blfootnote[1]{%
\begingroup
\renewcommand\thefootnote{}\footnote{#1}%
\addtocounter{footnote}{-1}%
\endgroup
}

\begin{document}

\title{3D Human Action Representation Learning via Cross-View Consistency Pursuit}

\author{
Linguo Li$^{1,2*}$\ \
Minsi Wang$^{1,2*}$\ \
Bingbing Ni$^{1,2**}$\ \
Hang Wang$^{1,2}$\ \
Jiancheng Yang$^{1,2}$\ \
Wenjun Zhang$^{1}$\\
$^1$Shanghai Jiao Tong University, Shanghai 200240, China\\
$^2$MoE Key Lab of Artificial Intelligence, AI Institute, Shanghai Jiao Tong University\\
{\tt\small mswang1994@gmail.com, \{LLG440982, nibingbing, zhangwenjun\}@sjtu.edu.cn}
}

\maketitle
\pagestyle{empty}
\thispagestyle{empty}

\begin{abstract}
In this work, we propose a Cross-view Contrastive Learning framework for unsupervised 3D skeleton-based action Representation (CrosSCLR), by leveraging multi-view complementary supervision signal.
CrosSCLR consists of both single-view contrastive learning (SkeletonCLR) and cross-view consistent knowledge mining (CVC-KM) modules, integrated in a collaborative learning manner.
It is noted that CVC-KM works in such a way that high-confidence positive/negative samples and their distributions are exchanged among views according to their embedding similarity, ensuring cross-view consistency in terms of contrastive context, i.e., similar distributions.
Extensive experiments show that CrosSCLR achieves remarkable action recognition results on NTU-60 and NTU-120 datasets under unsupervised settings, with observed higher-quality action representations.
Our code is available at \url{https://github.com/LinguoLi/CrosSCLR}.
\end{abstract}

\section{Introduction}
\label{sec:intro}

\blfootnote{$^*$\ Equal Contribution}
\blfootnote{$^{**}$\ Corresponding Author: Bingbing Ni}
\vspace{-10pt}

Human action recognition is an important but challenging task in computer vision research.
Due to the light-weight and robust estimation algorithms~\cite{cao2019openpose, xu2020deep},
3D skeleton has become a popular feature representation to study human action dynamics.
Many 3D action recognition works ~\cite{du2015hierarchical, zhang2017view, ke2017new, liang2019three, si2019attention, liu2020disentangling, kay2017kinetics} use a fully-supervised manner and require massive labeled 3D skeleton data.
However, annotating data is expensive and time-consuming, which prompts people to explore unsupervised methods~\cite{zheng2018unsupervised, lin2020ms2l, rao2020augmented, su2020predict} on skeleton data.
Some unsupervised methods exploit structure completeness within each sample based on pretext tasks, including reconstruction~\cite{gui2018adversarial, zheng2018unsupervised}, auto-regression~\cite{kundu2019unsupervised, su2020predict} and jigsaw puzzles~\cite{noroozi2016unsupervised, wei2019iterative}, but it is unsure that the designed pretext tasks generalize well for downstream tasks.
Other unsupervised methods are based on contrastive learning~\cite{wu2018unsupervised, chen2020simple, he2020momentum, khosla2020supervised}, aiming to leverage the instance discrimination of samples in latent space.

\begin{figure}[t]
\begin{center}
  \includegraphics[width=\linewidth]{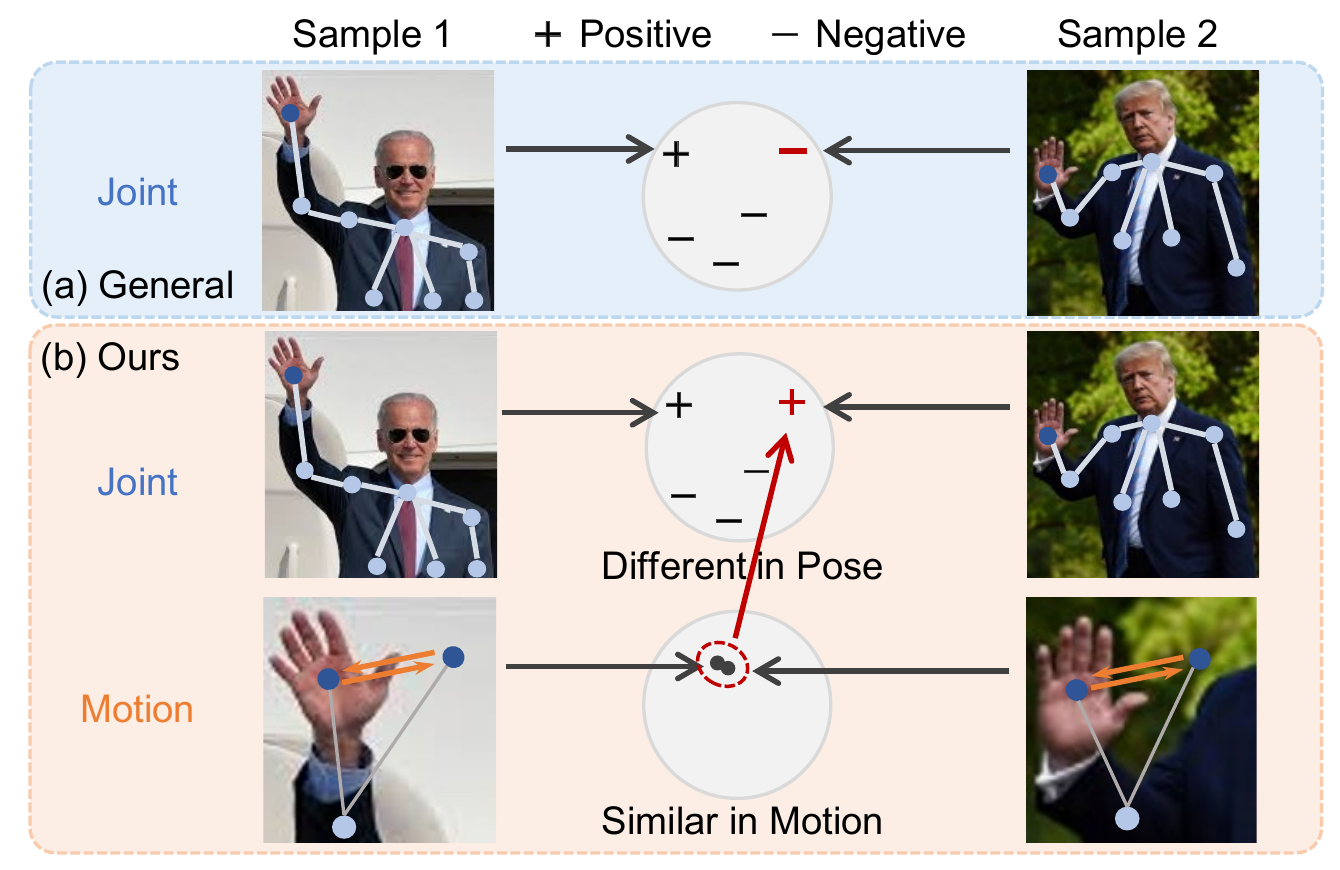}
\end{center}
\vspace{-5pt}
\caption{Hand waving in joint and motion form.
Two samples are from the same action class.
(a) usual contrastive learning methods regard them as negative pairs.
(b) in a multi-view situation, considering their similar motion patterns, they can be positive pairs.
This motivates us to introduce cross-view contrastive learning for skeleton representation.
}
\vspace{-10pt}
\label{fig:intro}
\end{figure}

Although the above approaches improve the skeleton representation capability to some extent, it is believed that the power of unsupervised methods is by far from fully explored.
On the one hand, traditional contrastive learning uses only one positive pair generated by data augmentation and even similar samples are regarded as negative samples.
Despite the high similarity, the negative samples are forced away in embedding space, which is unreasonable for clustering.
On the other hand, current unsupervised methods~\cite{ben2018coding, zheng2018unsupervised, kundu2019unsupervised, lin2020ms2l, su2020predict} have not yet explored the rich intra-supervision information provided by different skeleton modalities.
Considering that it is easy to obtain skeleton data in multiple ``views'', e.g., joint, motion and bone, complementary information preserved in different views can assist the operation to mine positive pairs from similar negative samples.
As shown in \figLabel\ref{fig:intro}, the same hand waving actions are different in pose (joint), but similar in motion.
Usual contrastive learning methods regard them as negative pairs, keeping them away in embedding space.
If such complementary information, i.e., different in joint but similar in motion, could be fully utilized and explored, the size of hidden positive pairs in joint can be boosted, enhancing training fidelity.
Thus, the cross-view contrastive learning strategy takes advantage of multi-view knowledge, resulting in better-extracted skeleton features.

To this end, we propose a Cross-view Contrastive Learning framework for Skeleton-based action Representation (CrosSCLR), which exploits multi-view information for mining positive samples and pursuing cross-view consistency in unsupervised contrastive learning, enabling the model to extract more comprehensive cross-view features.
First, parallel Contrastive Learning is evoked for each single-view Skeleton action Representation (SkeletonCLR), yielding multiple single-view embedding features.
Second, inspired by the fact that the distance of samples in embedding space reflects the similarity of the samples in the original space, we refer to the extreme similarity of samples in one view to guide the learning process in another view, as shown in \figLabel\ref{fig:intro}.
More specifically, Cross-View Consistent Knowledge Mining (CVC-KM) module is developed to exam the similarity of samples, and select the most similar pairs as positive ones to boost the positive set in complementary views, i.e., \textit{embedding distance/similarity (confidence score) serves as the weight of corresponding mined sample in the contrastive loss}.
In other words, CVC-KM conveys the most prominent knowledge from one view to others, introduces complementary pseudo-supervised constraint and promotes information sharing among views.
The entire framework excavates positive pairs across views according to the distance between samples in the embedding space to promote knowledge exchange among views, so that the extracted skeleton features will contain multi-view knowledge and are more competitive for various downstream tasks.
Extensive results on NTU-RGB+D~\cite{shahroudy2016ntu, liu2019ntu} datasets demonstrate that our method indeed boosts the 3D action representation learning benefiting from cross-view consistency.
We summarize our contributions as follows:

\begin{itemize}[topsep=0pt, partopsep=5pt, leftmargin=13pt, parsep=0pt, itemsep=4pt]
\item We propose CrosSCLR, a cross-view contrastive learning framework for skeleton-based action representation.
\item We develop Contrastive Learning for Skeleton-based action Representation (SkeletonCLR) to learn the single-view representations of skeleton data.
\item We use parallel SkeletonCLR models and CVC-KM to excavate useful samples across views, enabling the model to capture more comprehensive representation unsupervisedly.
\item We evaluate our model on 3D skeleton datasets, e.g., NTU-RGB+D 60/120, and achieve remarkable results under unsupervised settings.
\end{itemize}

\section{Related Work}
\mysection{Self-Supervised Representation Learning}
Self-supervised learning is to learn feature representations from numerous unlabeled data, which usually generates supervision by pretext tasks, e.g., jigsaw puzzles~\cite{noroozi2016unsupervised, noroozi2018boosting, wei2019iterative}, colorization~\cite{zhang2016colorful}, predicting rotation~\cite{gidaris2018unsupervised, zhai2019s4l}.
For sequence data, supervision can be generated by frame orders~\cite{misra2016shuffle, fernando2017self, lee2017unsupervised}, space-time cubic puzzles~\cite{kim2019self} and prediction~\cite{wang2019self, lin2020ms2l}, but these methods highly rely on the quality of pretext tasks.
Recently, contrastive methods~\cite{oord2018representation, wu2018unsupervised, tian2019contrastive, khosla2020supervised} based on instance discrimination have been proposed for representation learning.
MoCo~\cite{he2020momentum} introduces a memory bank to store the embeddings of negative samples, and SimCLR~\cite{chen2020simple} uses a much larger mini-batch size to compute the embeddings in real time, but they can not capture the cross-view knowledge for 3D action representation.
Concurrent work CoCLR~\cite{han2020self} leverages multi-modal information for video representation via co-training, which doesn't consider the contrastive context.
Our CrosSCLR simultaneously trains models in all views by encouraging cross-view consistency, leading to more representative embeddings.

\mysection{Skeleton-based Action Recognition}
To tackle skeleton-based action recognition tasks, early approaches are generally based on hand-craft features~\cite{ni2011rgbd, wang2012mining, vemulapalli2014human, vemulapalli2016rolling}.
Recent methods pay more attention to deep neural networks.
For the sequence structure of skeleton data, many RNN-based methods~\cite{du2015hierarchical, shahroudy2016ntu, zhang2017view, song2018spatio, zhang2019view} were carried out to effectively utilize the temporal feature.
Since RNN suffers from gradient vanishing~\cite{hochreiter2001gradient}, CNN-based models~\cite{ke2017new, li2017skeleton} attract researchers' attention, but they need to convert skeleton data to another form.
Further, ST-GCN~\cite{stgcn2018aaai} was proposed to better model the graph structure of skeleton data. Then the attention mechanism~\cite{li2019actional, shi2019two, si2019attention, zhang2020context, zhang2020semantics} and multi-stream structure~\cite{liang2019three, shi2019skeleton, shi2019two, wang2020learning} are applied to adaptively capture multi-stream features based on GCNs.
We adopt the widely-used ST-GCN as the backbone to extract the skeleton features.

\mysection{Unsupervised Skeleton Representation}
Many unsupervised methods~\cite{srivastava2015unsupervised, luo2017unsupervised, martinez2017human, li2018unsupervised} were proposed to capture action representations in videos.
For skeleton data, previous works~\cite{zanfir2013moving, ben2018coding} have achieved some progress in unsupervised representation learning without deep neural networks.
Recent deep learning methods~\cite{gui2018adversarial, zheng2018unsupervised, kundu2019unsupervised, su2020predict} are based on the structure of encoder-decoder or generative adversarial network (GAN).
LongT GAN~\cite{zheng2018unsupervised} proposed an auto-encoder-based GAN for sequential reconstruction and evaluated it on the action recognition tasks.
P\&C~\cite{su2020predict} uses a weak decoder in the encoder-decoder model, forcing the encoder to learn more discriminative features.
MS$^2$L~\cite{lin2020ms2l} proposed a multi-task learning scheme for action representation learning.
However, these methods highly depend on reconstruction or prediction,
and they do not exploit the natural multi-view knowledge of skeleton data.
Thus, we introduce CrosSCLR for unsupervised 3D action representation.

\section{CrosSCLR}

Although 3D skeleton has shown its importance in action recognition, unsupervised skeleton representation has not been well exploited recently. Since the easily-obtained “multi-view” skeleton information plays a significant role in action recognition, we expect to exploit them to mine positive samples and pursue cross-view consistency in unsupervised contrastive learning, thus giving rise to a Cross-view Contrastive Learning (CrosSCLR) framework for Skeleton-based action Representation.

\begin{figure*}[t]
\centerline{\includegraphics[width=\textwidth, trim=0 0 0 0]{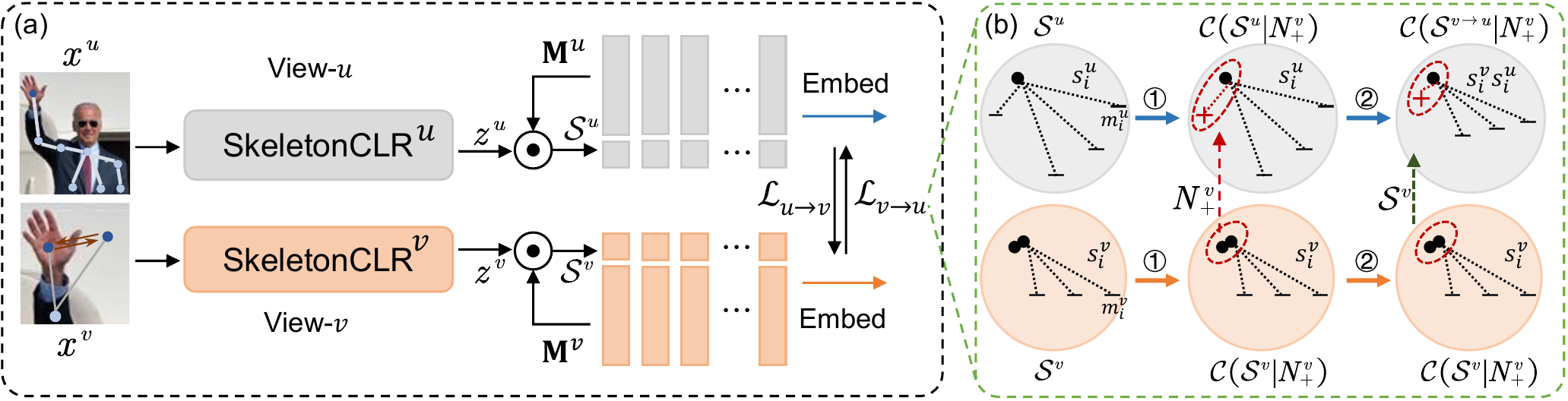}}
\caption{(a) CrosSCLR. Given two samples $x^u, x^v$ generated from the same raw data, e.g., joint and motion, SkeletonCLR models produce single-view embeddings while cross-view consistent knowledge mining (CVC-KM) exchanges multi-view complementary knowledge.
(b) how $\mathcal{L}_{v\rightarrow u}$ works in embedding space. In step 1, we mine high-confidence knowledge $N^v_+$ from similarities $\mathcal{S}^v$ to boost the positive set of view $u$, i.e., $z^u$ shares $z^v$'s neighbors; In step 2, we use the similarities $\mathcal{S}^v$ to supervise the embedding distribution in view $u$. $z^u, z^v$ share similar relationships with others. Thus, two embedding spaces become similar under the constraint of $\mathcal{L}_\text{cross}$.
}
\label{fig:struc}
\vspace{-10pt}
\end{figure*}

As shown in \figLabel\ref{fig:struc}, CrosSCLR contains two key modules: 1) SkeletonCLR (\secLabel\ref{sec:moco}): a contrastive learning framework to unsupervisedly learn single-view representations, and 2) CVC-KM (\secLabel\ref{sec:cro}): it conveys the most prominent knowledge from one view to others, introduces complementary pseudo-supervised constraint and promotes information sharing among views.
Finally, the more discriminating representations can be obtained by cooperatively training (\secLabel\ref{sec:cro}).

\subsection{Single-View 3D Action Representation}
\label{sec:moco}
Contrastive learning has been widely-used due to its instance discrimination capability, especially for images~\cite{chen2020simple, he2020momentum} and videos~\cite{han2020self}.
Inspired by this, we develop \textit{SkeletonCLR} to learn single-view 3D action representations, based on the recent advanced practice, MoCo~\cite{he2020momentum}.

\begin{figure}[t]
	\begin{center}
	  \includegraphics[width=\linewidth, trim=0 20 0 20]{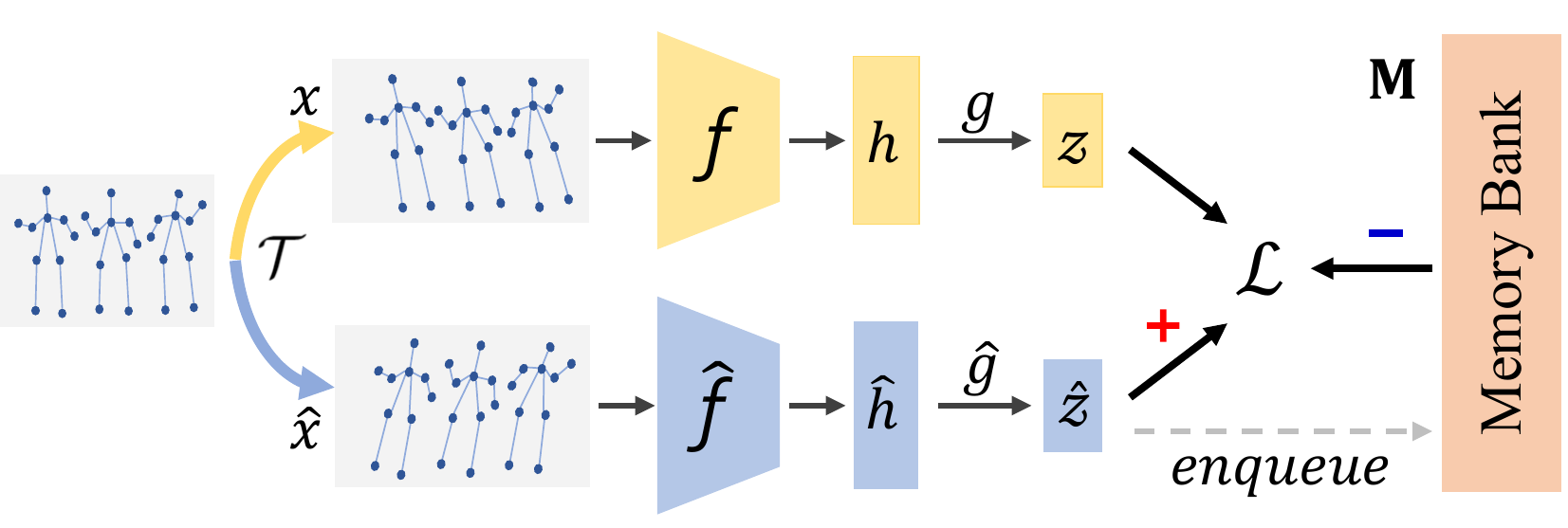}
	\end{center}
\caption{Architecture of single-view SkeletonCLR, which is a memory augmented contrastive learning framework.}
\label{fig:moco}
\vspace{-10pt}
\end{figure}

\mysection{SkeletonCLR}
It is a memory-augmented contrastive learning method for skeleton representation, which considers one sample's different augments as its positive samples and other samples as negative samples.
In each training step, the batch embeddings are stored in first-in-first-out memory to get rid of redundant computation, serving as negative samples for the next steps.
The positive samples are embedded close to each other while the embeddings of negative samples are pushed away.
As shown in \figLabel\ref{fig:moco}, SkeletonCLR consists of the following major components:

\begin{itemize}[topsep=0pt, partopsep=5pt, leftmargin=13pt, parsep=0pt, itemsep=4pt]
\item A data augmentation module $\mathcal{T}$ that randomly transforms the given skeleton sequence into different augments $x$, $\hat{x}$ that are considered as positive pairs.
For skeleton data, we adopt \textit{Shear} and \textit{Crop} as the augmentation strategy (see \secLabel\ref{sec:md} and Appendix).

\item Two encoders ${f}$ and $\hat{f}$ that embed ${x}$ and $\hat{x}$ into hidden space: ${h}={f}({x};{\theta})$ and $\hat{h}=\hat{f}(\hat{x};\hat{\theta})$, where ${h}, \hat{h} \in \mathbb{R}^{c_h}$.
$\hat{f}$ is the momentum updated version of ${f}$: $\hat{\theta}\leftarrow \alpha \hat{\theta}+(1-\alpha){\theta}$,
where $\alpha$ is a momentum coefficient.
SkeletonCLR uses ST-GCN~\cite{stgcn2018aaai} as the backbone (Details are in \secLabel\ref{sec:md}).

\item A simple projector ${g}$ and its momentum updated version $\hat{g}$ that project the hidden vector to a lower dimension space: ${z}={g}({h})$, $\hat{z}=\hat{g}(\hat{h})$, where ${z}, \hat{z} \in \mathbb{R}^{c_z}$.
The projector is a fully-connected (FC) layer with ReLU.

\item A memory bank $\mathbf{M}=\{m_i\}_{i=1}^M$ that stores negative samples to avoid redundant computation of the embeddings.
It is a first-in-first-out queue updated per iteration by $\hat{z}$.
After each inference step, $\hat{z}$ will enqueue while the earliest embedding in $\mathbf{M}$ will dequeue.
During contrastive training, $\mathbf{M}$ provides numerous negative embeddings while the new calculated $\hat{z}$ is the positive embedding.

\item An InfoNCE~\cite{oord2018representation} loss for instance discrimination:
\begin{align}
\label{eq:Linfo}
\mathcal{L}&=-\log\frac{\exp({z}\cdot \hat{z}/\tau)}{\exp({z}\cdot \hat{z}/\tau)+\sum^{M}_{i=1}{\exp({z}\cdot m_i/\tau)}}
\end{align}
where $m_i\in\mathbf{M}$, $\tau$ is the temperature hyper-parameter~\cite{hinton2015distilling}, and dot product ${z}\cdot \hat{z}$ is to compute their similarity where ${z}, \hat{z}$ are normalized.
\end{itemize}
Constrained by contrastive loss $\mathcal{L}$, the model is unsupervisedly trained to discriminate each sample in the training set.
At last, we can obtain a strong encoder ${f}$ that is beneficial to extract single-view distinguishing representations.

\mysection{Limitations of Single-View Contrastive Learning}
The above SkeletonCLR still suffers the following limitations:

1) \textit{Embedding distribution can provide more reliable information.}
We expect samples from the same category are embedded closely.
However, instance discrimination in SkeletonCLR uses only one positive pair and even similar samples are regarded as negative samples.
It is unreasonable that the negative samples are forced away in embedding space despite their high embedding similarity.
In other words, one positive pair cannot fully describe the relationships of samples, and a more reliable embedding distribution is needed, i,e., positive/negative setting plus embedding similarity.
We aim to mine more representative knowledge to facilitate contrastive learning, which is also the knowledge we want to exchange across views.
Thus, we introduce the \textit{contrastive context} in \secLabel\ref{sec:cro}.

2) \textit{Multi-view data can benefit representation learning.}
SkeletonCLR only relies on single-view data.
As shown in \figLabel\ref{fig:intro}, since we don't have any annotations, different samples of the same class are inevitably embedded into distinct places far from each other, i.e., they distribute sparsely/irregularly, bringing much difficulty for linear classification.
Considering the readily generated multi-view data of 3D skeleton (see \secLabel\ref{sec:md}), if such complementary information in \figLabel\ref{fig:intro}, i.e., different in joint but similar in motion, could be fully utilized and {explored}, the size of hidden positive pairs in joint can be boosted, enhancing training fidelity.
To this end, we inject this consideration into unsupervised contrastive learning framework.

\subsection{Cross-View Consistent Knowledge Mining}
\label{sec:cro}
Motivated by the situation in \figLabel\ref{fig:intro} that complementary knowledge is preserved in multiple views, we propose the Cross-View Consistent Knowledge Mining (CKC-KM), leveraging the high similarity of samples in one view to guide the learning process in another view.
It excavates positive pairs across views according to the embedding similarity to promote knowledge exchange among views, then the size of hidden positive pairs in each view can be boosted and the extracted skeleton features will contain multi-view knowledge, resulting in a more regular embedding space.

In this section, we first clarify \textit{contrastive context} as the consistent knowledge across views, and then show how to mine high-confidence knowledge, and finally inject its cross-view consistency into single-view SkeletonCLR to further benefit the cross-view unsupervised representation.

\mysection{Contrastive Context as Consistent Knowledge}
As discussed above, the knowledge we want to exchange across views is one sample's \textit{contrastive context}, which describes this sample's relationships with others (distribution) in embedding space under the settings of contrastive learning.
Notice that SkeletonCLR uses a memory bank to store necessary embeddings.
Given one sample's embedding $z$ and corresponding memory bank $\mathbf{M}$, its {contrastive context} is a similarity set $\mathcal{S}$ among $z$ and $\mathbf{M}$ conditioned on specific knowledge miner $\Gamma$ that generates index set $N_+$ of positive samples, 
\begin{align}
\mathcal{S} &= \{s_i\}_{i\in N} = \{z\cdot m_i\}_{i\in N} \\
(\mathcal{S}_+, N_+) &= \Gamma(\mathcal{S}) 
\end{align}
where $\mathcal{S}_{+} = \{s_i\}_{i\in N_+}$ and dot product ``$\cdot$'' is to compute the similarity $s_i$ among embeddings $z$ and $m_i$. $N$ is the index set of embeddings in memory bank and $N_+$ is the index set of positive samples selected by knowledge miner $\Gamma$.
Thus contrastive context $\mathcal{C}(\mathcal{S}| N_+)$ consists of following two aspects:

\begin{itemize}[topsep=0pt, partopsep=5pt, leftmargin=13pt, parsep=0pt, itemsep=4pt]
	\item \textit{Embedding Context} $\mathcal{S}$: it is the relationship between one sample and others in embedding space, i.e., distribution;
	\item \textit{Contrastive Setting} $N_+$: it is the positive setting mined by $\Gamma$ according to the embedding similarity $\mathcal{S}$;
\end{itemize}
thus $\mathcal{C}(\mathcal{S}| N_+)=\{\mathcal{S}_{+}, \mathcal{S}_{-}\}$ has \textit{positive} context $\mathcal{S}_{+}$ and \textit{negative} context $\mathcal{S}_{-}$, where $\mathcal{S} = \mathcal{S}_{+} \cup \mathcal{S}_{-}$.
The contrastive context contains not only the information of the most similar samples but the detailed relationships of samples (distribution). 

In \eqLabel(\ref{eq:Linfo}), the embedding $z$ has positive context $S_{+}=\{z\cdot \hat{z}\}$ which does not consider any of neighbors in embedding space except for the augments.
Despite the high similarity, the negative samples are forced away in embedding space, and then samples belonging to the same category are difficultly embedded into the same cluster, which is not efficient to build a ``regular'' embedding space for down-stream classification tasks.

\mysection{High-confidence Knowledge Mining}
To solve the above issue, we develop the high-confidence Knowledge Mining mechanism (KM), which selects the most similar pairs as positive ones to boost the positive sets.
It shares similar high-level spirit with neighborhood embedding~\cite{hinton2002stochastic} but performs differently in an unsupervised contrastive manner.

Specifically, it is based on the following observation in \figLabel\ref{fig:tsne} that {after single-view contrastive learning, two embeddings most likely belong to the same category if they are embedded closely enough; on the contrary, two embeddings hardly belong to the same class if they locate extremely far from each other in embedding space}.
Therefore, we can facilitate contrastive learning by setting the most similar embeddings as positive to make it more clustered: 
\begin{align}
	\Gamma(\mathcal{S}) &= \text{Topk}(\mathcal{S}) \\
    \mathcal{L}_\text{KM}&=-\log\frac{\exp({z}\cdot \hat{z}/\tau) + \sum_{i\in N_+}{\exp({z} \cdot m_i/\tau)}}{\exp({z}\cdot \hat{z}/\tau) + \sum_{i\in N}{\exp({z}\cdot m_i/\tau)}} \label{eq:km}
\end{align}
where $\Gamma=\text{Topk}$ is the function to select the index of top-$K$ similar embeddings and $N_+$ is their index set in memory bank.
Compared to \eqLabel(\ref{eq:Linfo}), \eqLabel(\ref{eq:km}) will leads to a more regular space by pulling close more high-confidence positive samples.
Additionally, since we don't have any labels, a larger $K$ may harm the contrastive performance (see \secLabel\ref{sec:as}). 

\mysection{Cross-View Consistency Learning}
Considering easily-obtained multi-view skeleton data, complementary information preserved in different views can assist the operation to mine positive pairs from similar negative samples in \figLabel\ref{fig:intro}.
Then the size of hidden positive pairs can be boosted by cross-view knowledge communication, resulting in better-extracted skeleton features.
To this end, we design the cross-view consistency learning which not only mines the high-confidence positive samples from complementary view but also lets the embedding context be consistent in multiple views.
Its two-view case is illustrated in \figLabel\ref{fig:struc} for example.

Specifically, samples $x^u$ and $x^v$ are generated from the same raw data $x$ by the view generation method in \secLabel\ref{sec:md}, where $u$ and $v$ indicate two types of data views.
After single-view contrastive learning, two SkeletonCLR modules obtain the embeddings $z^u,z^v$ and corresponding memory bank $\mathbf{M}^u,\mathbf{M}^v$ respectively.
We can mine the high-confidence knowledge from two views by $(\mathcal{S}^u_{+}, N^u_+) = \text{Topk}(\mathcal{S}^u)$ and $(\mathcal{S}^v_{+}, N^v_+) = \text{Topk}(\mathcal{S}^v)$, where $\mathcal{S}^u_{+}, \mathcal{S}^v_{+}$ are the positive context of $z^u,z^v$ respectively.

CrosSCLR aims to learn the consistent embedding distribution in different views by encouraging the similarity of contrastive context, i.e., exchanging high-confidence knowledge across views.
In \figLabel\ref{fig:struc} (b), if we want to use the knowledge of view $v$ to guide view $u$'s contrastive learning, it contains two aspects:
1) step 1: we select the most similar pairs (positive) in view $v$ as the positive sets in view $u$, i.e., $\mathcal{S}^u, N^v_+ \rightarrow \mathcal{C}(\mathcal{S}^u|N^v_+)$. Thus the sample $z^u$ shares $z^v$'s positive neighbors;
2) step 2: we use the embedding similarity $s^v_i = z^v\cdot m^v_i$ in view $v$ as the weight of corresponding embedding {$m^u_i$} in view $u$ to provide the detailed relational information, i.e., 
$m^{v\rightarrow u}_i = s^v_i m^u_i$.
Then the similarity is computed by $z^u\cdot m^{v\rightarrow u} = s^v_i(z^u\cdot m^u_i) =  s^u_i s^v_i$ and $z^u$ has embedding context $\mathcal{S}^{v\rightarrow u}= \{z^u\cdot m^{v\rightarrow u}_i  \}_{i \in N} = \{ s^u_i s^v_i\}_{i \in N}$.
Finally,
the overall loss is conducted as:
\begin{align}
    \mathcal{L}_{v\rightarrow u}&= -\log\frac{\exp({z^u}\cdot \hat{z^u}/\tau) + \sum_{i\in N^v_+}{\exp(s^u_i s^v_i)/\tau)}}{\exp({z^u}\cdot \hat{z^u}/\tau) + \sum_{i\in N}{\exp(s^u_i s^v_i)/\tau)}} \label{eq:Lvu} \\
    \mathcal{L}_\text{cross} &= \mathcal{L}_{u\rightarrow v} + \mathcal{L}_{v\rightarrow u} \label{eq:cross}
\end{align}
where $\mathcal{L}_{v\rightarrow u}$ means we transfer the contrastive context of $z^v$ to that of $z^u$.
Since $s^u_i, s^v_i$ are the embedding context of $z^u_i, z^v_i$, we call \eqLabel(\ref{eq:cross}) as \textit{cross-view contrastive context learning}, which constrains the similar distribution of two views (see \secLabel\ref{sec:as}, the results of t-SNE). 
Compared to \eqLabel(\ref{eq:km}), \eqLabel(\ref{eq:Lvu}) considers the cross-view information, cooperatively using one view's high-confidence positive samples and its distribution to instruct {the other view's contrastive learning, resulting in more regular space and better extracted skeleton features.}

\mysection{Learning CrosSCLR}
For more views, CrosSCLR has following objective:
\begin{align}
    \mathcal{L}_\text{cross}&=\sum^{U}_{u}\sum^{U}_{v} \mathcal{L}_{u\rightarrow v}
\label{eq:Lcro}
\end{align}
where $U$ is the number of views and $v \neq u$.

In the early training process, the model is not stable and strong enough to provide reliable cross-view knowledge without the supervision of labels.
As the unreliable information may lead astray, it is not encouraged to enable cross-view communication too early.
We perform two-stage training for CrosSCLR:
1) each view of the model is individually trained with \eqLabel(\ref{eq:Linfo}) without cross-view communication.
2) then the model can supply high-confidence knowledge, so the loss function is replaced with \eqLabel(\ref{eq:Lcro}), starting cross-view knowledge mining.

\subsection{Model Details}
\label{sec:md}

\mysection{View Generation of 3D Skeleton}
Generally, 3D human skeleton sequence has $T$ frames with $V$ joints, and each joint has $C=3$ coordinate feature, which can be noted as $x\in\mathbb{R}^{C\times T\times V}$.
Different from videos, the views~\cite{shi2019skeleton, shi2019two} of skeleton, e.g., joint, motion, bone, motion of bone, can be easily obtained, which is a natural advantage for skeleton-based representation learning.
Motion is represented as the temporal displacement between frames: $x_{:,t+1,:} - x_{:,t,:}$, and bone is the distance between two neighboring joints in the same frame: $x_{:,:,v_2} - x_{:,:,v_1}$.
For simplicity, we use three views: joint, motion and bone in experiments.

\mysection{Encoder $f$}
We adopt ST-GCN~\cite{stgcn2018aaai} as encoder,
which is suitable for modeling graph-structure skeleton data by exploiting the spatial and temporal relations.
After a series of ST-GCN blocks, the output feature $x_{out}\in\mathbb{R}^{c_{out}\times t_{out}\times V}$ is applied by an average pooling operation on spatial and temporal dimensions, obtaining final representation $h\in\mathbb{R}^{c_h}$.

\section{Experiments}

\subsection{Datasets}

\mysection{NTU-RGB+D 60}
NTU-RGB+D 60 (NTU-60) dataset~\cite{shahroudy2016ntu} is a large-scale dataset of 3D joint coordinate sequences for skeleton-based action recognition, containing $56,578$ skeleton sequences in $60$ action categories.
Each skeleton graph contains $V=25$ body joints as nodes, and their 3D coordinates are initial features.
There are two protocols~\cite{shahroudy2016ntu} recommended.
1) Cross-Subject (xsub): training data and validation data are collected from different subjects.
2) Cross-View (xview): training data and validation data are collected from different camera views.

\mysection{NTU-RGB+D 120}
NTU-RGB+D 120 (NTU-120) dataset~\cite{liu2019ntu} is an extended version of NTU-60, containing $113,945$ skeleton sequences in $120$ action categories.
Two protocols~\cite{liu2019ntu} are recommended.
1) Cross-Subject (xsub): training data and validation data are collected from different subjects.
2) Cross-Setup (xset): training data and validation data are collected from different setup IDs.

\mysection{NTU-RGB+D 61-120}
NTU-RGB+D 61-120 (NTU-61-120) dataset is a subset of NTU-120 dataset, containing $57,367$ skeleton sequences in the last $60$ action categories in NTU-120.
The categories in NTU-61-120 do not intersect with those in NTU-60.
This dataset is used as external dataset to evaluate the transfer capability of our method.

\subsection{Experimental Settings}
\label{sec:es}

All the experiments are conducted on the PyTorch~\cite{paszke2017automatic} framework.
For data pre-processing, we remove the invalid frames of each skeleton sequence and then resize them to the length of $50$ frames by linear interpolation.
For optimization, we use SGD with momentum ($0.9$) and weight decay ($0.0001$).
The mini-batch size is set to $128$.

\mysection{Data Augmentation $\mathcal{T}$}
For skeleton sequence, we choose Shear~\cite{rao2020augmented} and Crop~\cite{shorten2019survey} as the augmentation strategy.

\textit{Shear} is a linear transformation on the spatial dimension.
The transformation matrix is defined as:
\begin{equation}
\small
    \mathbf{A}=\left[
    \begin{array}{ccc}
    1       &   a_{12}  &   a_{13}  \\
    a_{21}  &   1       &   a_{23}  \\
    a_{31}  &   a_{32}  &   1
    \end{array}
    \right]
\end{equation}
where $a_{12}$, $a_{13}$, $a_{21}$, $a_{23}$, $a_{31}$, $a_{32}$ are shear factors randomly sampled from $[-\beta,\beta]$.
$\beta$ is the shear amplitude.
The sequence $x$ is multiplied by the transformation matrix $\mathbf{A}$ on the channel dimension.
Then, the human pose in 3D coordinate is inclined at a random angle.

\textit{Crop} is an augmentation on the temporal dimension that symmetrically pads some frames to the sequence and then randomly crops it to the original length.
The padding length is defined as $T/\gamma$, $\gamma$ is noted as padding ratio.
The padding operation uses the reflection of the original boundary.

\begin{table}[t]
	\begin{center}
		\small
		\begin{tabular}{@{}llcccc@{}}
			\toprule
			&                        & \multicolumn{2}{c}{NTU-60 (\%)} \\
			Method      &View                    & xsub    & xview   \\ \midrule
			SkeletonCLR &Joint                   & 68.3   & 76.4   \\
			SkeletonCLR &Motion                  & 53.3   & 50.8   \\
			SkeletonCLR &Bone                    & 69.4   & 67.4   \\
			2s-SkeletonCLR &Joint + Motion          & 70.5   & 77.9   \\
			3s-SkeletonCLR &Joint + Motion + Bone   & 75.0   & 79.8   \\ \midrule
			CrosSCLR    &Joint                   & 72.9   & 79.9   \\
			CrosSCLR    &Motion                  & 72.7   & 77.6   \\
			CrosSCLR    &Bone                    & 75.2   & 78.8   \\
			2s-CrosSCLR    &Joint + Motion          & 74.5   & 82.1   \\
			3s-CrosSCLR    &Joint + Motion + Bone   & \textbf{77.8}   & \textbf{83.4}   \\ \bottomrule
		\end{tabular}
	\end{center}
	\vspace{-5pt}
	\caption{Comparisons of SkeletonCLR and CrosSCLR on each view and their ensembles. SkeletonCLR models are trained independently and ``+'' means the ensemble model.}
	\vspace{-10pt}
	\label{tbl:view}
\end{table}

\mysection{Unsupervised Pre-training}
We generate three views of skeleton sequences, i.e., joint, motion and bone.
For the encoder, we adopt ST-GCN~\cite{stgcn2018aaai}, but the number of channels in each layer is reduced to $1/4$ of the original setting.
For contrastive settings, we follow that in MOCOv2~\cite{chen2020improved} but reduce the size of memory bank $\mathbf{M}$ to $30$k.
For data augmentation, We set shear amplitude $\beta=0.5$ and the padding ratio $\gamma=6$.
The model is trained for $300$ epochs with the learning rate $0.1$ (multiplied by $0.1$ at epoch $250$).
InfoNCE loss in \eqLabel(\ref{eq:Linfo}) is used in the first $150$ epochs, and then replaced with $\mathcal{L}_{cross}$ in \eqLabel(\ref{eq:Lcro}) after $150$-th epoch.
We set $K=1$ as the default in the knowledge mining mechanism.

\mysection{Linear Evaluation Protocol}
The models are verified by linear evaluation for action recognition task, i.e., attaching the frozen encoder to a linear classifier (a \textit{fully-connected} layer followed by a \textit{softmax} layer), and then training the classifier supervisedly.
We train models for $100$ epochs with learning rate $3.0$ (multiplied by $0.1$ at epoch $80$).

\mysection{Finetune Protocol}
We append a linear classifier to the \textit{learnable} encoder, and then train the whole model for the action recognition task, to compare it with fully-supervised methods.
We train for $100$ epochs with learning rate $0.1$ (multiplied by $0.1$ at epoch $80$).

\subsection{Ablation Study}
\label{sec:as}
All experiments in this section are conducted on NTU-60 dataset and follow the unsupervised pre-training and linear evaluation protocol in \secLabel\ref{sec:es}.

\mysection{Effectiveness of CrosSCLR}
In \tblLabel\ref{tbl:view}, we separately pre-train SkeletonCLR and jointly pre-train CrosSCLR models on different skeleton views, e.g., joint, motion and bone.
We adopt linear evaluation on each view of the models.
\tblLabel\ref{tbl:view} reports that 1) CrosSCLR improves the capability of each single SkeletonCLR model, e.g., CrosSCLR-joint (79.88) \textit{v.s} SkeletonCLR-joint (76.44) on xview protocol;
2) CrosSCLR bridges the performance gap of two views and jointly improves their accuracy, e.g., for SkeletonCLR, joint (76.44) \textit{v.s} motion (50.82) but for CrosSCLR, joint (79.88) \textit{v.s} motion (77.59);
3) CrosSCLR improves the multi-view ensemble results via cross-view training.
In summary, the cross-view high-confidence knowledge does help the model extract more discriminating representations.

\begin{figure}[t]
\begin{center}
  \includegraphics[width=0.99\linewidth]{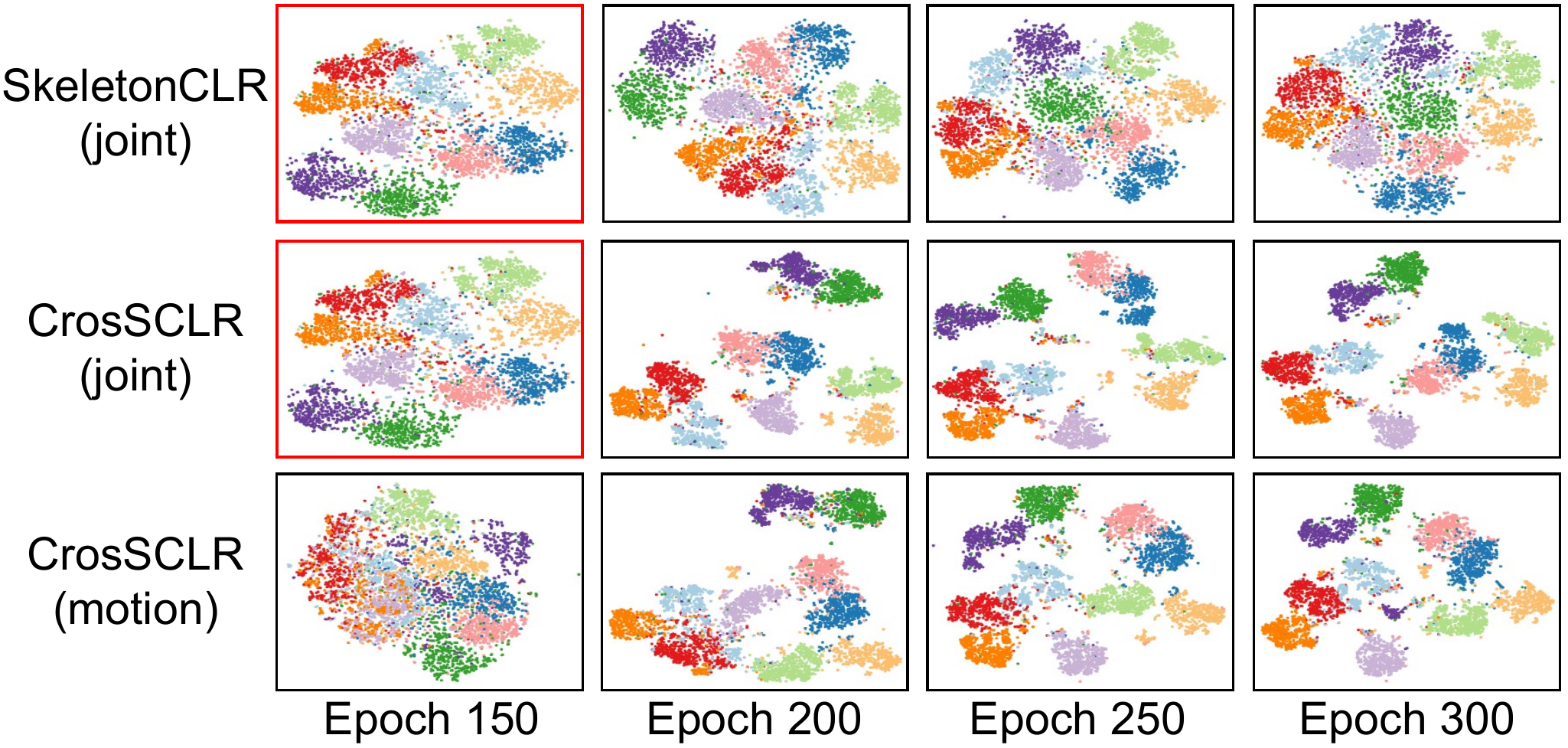}
\end{center}
\vspace{-5pt}
\caption{The t-SNE visualization of embeddings at different epochs during pre-training. Embeddings from $10$ categories are sampled and visualized with different colors. For CrosSCLR, $\mathcal{L}_\text{cross}$ starts to be available at epoch $150$, so its distribution has no difference from that of SkeletonCLR before epoch $150$, shown in red boxes.}
\vspace{-5pt}
\label{fig:tsne}
\end{figure}

\mysection{Qualitative Results}
We apply t-SNE~\cite{maaten2008visualizing} with fix settings to show the embedding distribution of SkeletonCLR and CrosSCLR on $150, 200, 250, 300$ epochs during pre-training in \figLabel\ref{fig:tsne}.
Note that cross-view loss, \eqLabel(\ref{eq:Lcro}), is available only after epoch $150$.
From the visual results, we can draw a similar conclusion to that in \tblLabel\ref{tbl:view}.
Embeddings of CrosSCLR are clustered more closely than that of SkeletonCLR, which is more discriminating.
For CrosSCLR, the distributions of joint and motion are distinct at $150$-th epoch but look very similar at $300$-th epoch, i.e., consistent distribution.
Especially, they both build a more ``regular'' space than SkeletonCLR, proving the effectiveness of CrosSCLR.

\mysection{Effects of Contrastive Setting top-$K$}
As hyper-parameter $K$ determines the number of mined samples, influencing the depth of knowledge exchange, we study how $K$ impacts the performance in cross-view learning.
\tblLabel\ref{tbl:K} shows that $K$ has a great influence on the performance and achieves the best result when $K=1$.
However, a larger $K$ decreases the performance, because the not so confident information may lead the model astray in an unsupervised case.

\mysection{Contrastive Setting $N_+$ and Embedding Context $\mathcal{S}$}
We develop following models in \tblLabel\ref{tbl:weight} for comparison:
1) SkeletonCLR + $\mathcal{L}_\text{KM}$ is a model with single-view knowledge mining.
2) CrosSCLR \textit{w/o.} embedding context (EC) is the model only using the contrastive setting $N_+$ for cross-view learning, which ignores the embedding context/distribution, i.e., $S^v_i=1, \forall i \in N$ in \eqLabel(\ref{eq:Lvu}).
The results of SkeletonCLR + $\mathcal{L}_\text{KM}$ show that KM improves the representation capability of SkeletonCLR.
Additionally, CrosSCLR achieves worse performance without using embedding context (EC), proving the significance of similarity/distribution among samples.

\mysection{Effects of Augmentations}
SkeletonCLR and CrosSCLR are based on contrastive learning, but the data augmentation strategy used on skeleton data is rarely explored, especially for the GCN encoder.
We verify the effectiveness of data augmentation and the impact of different augmented intensities in skeleton-based contrastive learning by conducting experiments on SkeletonCLR, as shown in \tblLabel\ref{tbl:aug}.
It indicates the importance of data augmentation in SkeletonCLR.
We choose $\beta=0.5$ and $\gamma=6$ as default settings according to the mean accuracy on xsub and xview protocols.

\begin{table}[t]
    \centering
	\parbox{.4\linewidth}{
		\centering
		\footnotesize
		\setlength{\tabcolsep}{6pt} 
        \begin{tabular}{@{}lccc@{}}
        \toprule
               & \multicolumn{2}{c}{NTU-60 (\%)} \\
         top-$K$  & xsub          & xview         \\ \midrule
         0        & 70.5          & 77.4          \\
         1        & \textbf{74.5} & \textbf{82.1} \\
         3        & 73.7          & 79.9          \\
         5        & 72.4          & 79.2          \\
         7        & 73.0          & 78.6          \\
         10       & 64.4          & 69.9          \\ \bottomrule
        \end{tabular}
        \vspace{5pt}
        \caption{Results of pre-training 2s-CrosSCLR with various $K$ in knowledge miner $\Gamma$. }
        \label{tbl:K}
		}\hfill
	\parbox{.55\linewidth}{
		\centering
		\footnotesize
		\setlength{\tabcolsep}{6pt} 
        \begin{tabular}{@{}cc|cc@{}}
        \toprule
        \multicolumn{2}{l}{Augmentation}& \multicolumn{2}{|c}{NTU-60 (\%)}   \\
        Shear $\beta$   & Crop $\gamma$ & xsub          & xview        \\ \midrule
        $0$             & $0$           & 33.3          & 26.2         \\
        $0.2$           & $0$           & 62.7          & 67.7         \\
        $0.5$           & $0$           & \textbf{66.3} &\textbf{68.8} \\
        $1.0$           & $0$           & 62.0          & 66.8         \\ \midrule
        $0.5$           & $4$           & 67.6          & 76.3         \\
        $0.5$           & $6$           & 68.3          & \textbf{76.4}\\
        $0.5$           & $8$           & \textbf{69.1} & 74.7         \\\\ 
        \bottomrule
        \end{tabular}
        \vspace{5pt}
        \caption{Ablation study on different data augmentations for SkeletonCLR (joint).}
        \label{tbl:aug}
	    }\hfill        
\end{table}

\begin{table}[]
\small
\begin{center}
\begin{tabular}{@{}llccc@{}}
\toprule
                                        & Views of       & \multicolumn{2}{c}{NTU-60 (\%)} \\
Method                                  & Pre-training   & xsub     & xview     \\ \midrule
SkeletonCLR                             & Joint          & 68.3     & 76.4     \\
SkeletonCLR + $\mathcal{L}_\text{KM}$   & Joint          & 69.3     & 77.4     \\
CrosSCLR \textit{w/o.} EC               & Joint + Motion & 71.4     & 78.5     \\
CrosSCLR                                & Joint + Motion & \textbf{72.9}     & \textbf{79.9}     \\ \bottomrule
\end{tabular}
\end{center}
\vspace{-5pt}
\caption{Ablation study on contrastive settings $N_+$ and embedding context (EC).
The models are linear evaluated on \textbf{only joint}.}
\vspace{-5pt}
\label{tbl:weight}
\end{table}

\subsection{Comparison}
We compare CrosSCLR with other methods under \textit{linear evaluation} and \textit{finetune} protocols.
Since the backbone in many methods is an RNN-based model, e.g., GRU or LSTM, we additionally use LSTM (following the setting in \cite{rao2020augmented}) as the encoder for a fair comparison, i.e., CrosSCLR (LSTM).

\mysection{Unsupervised Results on NTU-60}
In \tblLabel\ref{tbl:unsup60}, LongT GAN~\cite{zheng2018unsupervised} adversarially trains the model by skeleton inpainting pretext task, MS$^2$\text{L~\cite{lin2020ms2l}} trains the model by multi-task scheme, i.e, prediction, jiasaw puzzle and instance discrimination, AS-CAL~\cite{rao2020augmented} uses momentum LSTM encoder for contrastive learning with single-view skeleton sequence, P\&C~\cite{su2020predict} trains a stronger encoder by weakening decoder, and SeBiReNet~\cite{nie2020unsupervised} constructs a human-like GRU network to utilize view-independent and pose-independent feature.
{Our CrosSCLR exploits the multi-view knowledge by cross-view consistent knowledge mining.
Taking a fully-connected layer (FC) as the classifier, our model outperforms other methods with the same classifier.
With LSTM classifier and LSTM encoder, our model outperforms the above methods on both xsub and xview protocols.
}

{\mysection{Results on NTU-120}
As few unsupervised results are reported on NTU-120 dataset, we compare our method with unsupervised and supervised methods.
As shown in \tblLabel\ref{tbl:120}, TSRJI~\cite{caetano2019skeleton} supervisedly utilizes attention LSTM, AS-CAL~\cite{rao2020augmented} adopts LSTM for skeleton modeling, and our method defeats the other unsupervised method and some of the supervised methods.
}

\begin{table}[]
\begin{center}
\small
\begin{tabular}{@{}l|ll|cc@{}}
\toprule
                                      &             &             & \multicolumn{2}{c}{NTU-60 (\%)}\\
Method                                & Encoder    & Classifier  & xsub       & xview    \\ \midrule
LongT GAN~\cite{zheng2018unsupervised} & GRU         & FC          & 39.1       & 48.1     \\
MS$^2$\text{L~\cite{lin2020ms2l}}     & GRU         & GRU         & 52.6       & -        \\
AS-CAL~\cite{rao2020augmented}        & LSTM        & FC          & 58.5       & 64.8     \\
P\&C~\cite{su2020predict}             & GRU         & KNN         & 50.7       & 76.3     \\
SeBiReNet~\cite{nie2020unsupervised}  & GRU         & LSTM        & -          & 79.7     \\ \midrule
3s-CrosSCLR (LSTM)                       & LSTM        & FC          & 62.8       & 69.2     \\
3s-CrosSCLR (LSTM)                       & LSTM        & LSTM        & 70.4       & 79.9     \\
3s-CrosSCLR$\ddagger$                    & ST-GCN       & FC          & 72.8       & 80.7     \\
3s-CrosSCLR                              & ST-GCN       & FC          & \textbf{77.8}       & \textbf{83.4}     \\ \bottomrule
\end{tabular}
\end{center}
\vspace{-5pt}
\caption{Unsupervised results on NTU-60. These methods are pre-trained to learn encoder and then follow the linear evaluation protocol to learn the classifiers. ``$\ddagger$'' indicates the model pre-trained on NTU-61-120.}
\label{tbl:unsup60}
\end{table}

\begin{table}[]
\small
\begin{center}
\begin{tabular}{@{}lccc@{}}
\toprule
                                        &               & \multicolumn{2}{c}{NTU-120 (\%)} \\
Method                                  & Supervision   & xsub            & xset           \\ \midrule
Part-Aware LSTM~\cite{shahroudy2016ntu} & Supervised    & 25.5            & 26.3           \\
Soft RNN~\cite{hu2018early}             & Supervised    & 36.3            & 44.9           \\
TSRJI~\cite{caetano2019skeleton}        & Supervised    & 67.9            & 62.8           \\
ST-GCN~\cite{stgcn2018aaai}							        & Supervised    & \textbf{79.7}            & \textbf{81.3}      \\
\midrule
AS-CAL~\cite{rao2020augmented}          & Unsupervised  & 48.6            & 49.2           \\
3s-CrosSCLR (LSTM)                         & Unsupervised  & 53.9            & 53.2           \\
3s-CrosSCLR                                & Unsupervised  & \textbf{67.9}            & \textbf{66.7}           \\ \bottomrule
\end{tabular}
\end{center}
\vspace{-5pt}
\caption{Unsupervised results on NTU-120. We show and compare our method with unsupervised and supervised methods.}
\vspace{-5pt}
\label{tbl:120}
\end{table}

{
\mysection{Linear Classification with Fewer Labels}
We follow the same protocol as that of MS$^2$\text{L~\cite{lin2020ms2l}}, i.e., pre-training with all training data and then finetuning the classifier with only $1\%$ and $10\%$ randomly-selected labeled data respectively.
As shown in \tblLabel\ref{tbl:semisup}, CrosSCLR achieves higher performance than other methods.
}

\begin{table}[]
\small
\begin{center}
\begin{tabular}{@{}lccc@{}}
\toprule
                                        &                   & \multicolumn{2}{c}{NTU-60 (\%)}  \\
Method                                  & Label Fraction    & xsub           & xview           \\ \midrule
LongT GAN~\cite{zheng2018unsupervised}  & 1\%               & 35.2           & -               \\
MS$^2$\text{L~\cite{lin2020ms2l}}       & 1\%               & 33.1           & -               \\
3s-CrosSCLR                                & 1\%               & \textbf{51.1}           & \textbf{50.0}            \\ \midrule
LongT GAN~\cite{zheng2018unsupervised}  & 10\%              & 62.0           & -               \\
MS$^2$\text{L~\cite{lin2020ms2l}}       & 10\%              & 65.2           & -               \\
3s-CrosSCLR                                & 10\%              & \textbf{74.4}           & \textbf{77.8}            \\ \bottomrule
\end{tabular}
\end{center}
\vspace{-5pt}
\caption{Linear classification with fewer labels on NTU-60.}
\label{tbl:semisup}
\end{table}

\begin{table}[]
\small
\begin{center}
\begin{tabular}{@{}lcccc@{}}
\toprule
                            & \multicolumn{2}{c}{NTU-60 (\%)} & \multicolumn{2}{c}{NTU-120 (\%)} \\
Method                      & xsub      & xview     & xsub       & xset      \\ \midrule
3s-ST-GCN$^{*}$~\cite{stgcn2018aaai} & 85.2     & 91.4     & 77.2      & 77.1     \\ 
3s-CrosSCLR$\ddagger$ (FT)     & 85.6     & 92.0     & -          & -         \\
3s-CrosSCLR (FT)               & \textbf{86.2}     & \textbf{92.5}     & \textbf{80.5}      &\textbf{80.4}     \\ \bottomrule
\end{tabular}
\end{center}
\vspace{-5pt}
\caption{Finetuned results on NTU-60 and NTU-120. ST-GCN$^{*}$ is the method reproduced by released code. ``$\ddagger$'' indicates the model pre-trained on NTU-61-120. ``FT'' means finetune protocol.}
\vspace{-5pt}
\label{tbl:sup}
\end{table}

\mysection{Finetuned Results on NTU-60 and NTU-120}
We first unsupervisedly pre-train our model and follow the \textit{finetune} protocol for evaluation.
For fair comparison, ST-GCN$^*$~\cite{stgcn2018aaai} in \tblLabel\ref{tbl:sup} has the same number of parameters as 3s-CrosSCLR ($1/4$ channel with three streams).
It shows that the finetuned model, CrosSCLR (FT) outperforms the supervised ST-GCN on both NTU-60 and NTU-120 datasets, indicating the effectiveness of cross-view pre-training.

\mysection{Transfer Ability}
{We first pre-train CrosSCLR on NTU-61-120, and then transfer it to NTU-60 for linear evaluation, noted as CrosSCLR$\ddagger$.
The model trained under xsub protocol is transferred to the xsub protocol of NTU-60; the model trained under xset protocol is transferred to the xview protocol of NTU-60.
In \tblLabel\ref{tbl:unsup60}, it achieves better results than the other unsupervised methods, and its supervisedly finetuning result is higher than ST-GCN as shown in \tblLabel\ref{tbl:sup}.}

\section{Conclusion}
In this work, we propose a Cross-view Contrastive Learning framework for unsupervised 3D skeleton-based action representation to exploit multi-view high-confidence knowledge as complementary supervision.
It integrates single-view contrastive learning with cross-view consistent knowledge mining modules which convey the contrastive settings and embedding context among views by high-confidence sample mining.
Experiments show remarkable results of CrosSCLR for action recognition on NTU datasets.

\section*{Acknowledgement}
This work was supported by National Science Foundation of China (U20B2072, 61976137).
This work was also supported by NSFC (U19B2035), Shanghai Municipal Science and Technology Major Project (2021SHZDZX0102).

\clearpage

{\small
\bibliographystyle{ieee_fullname}
\bibliography{cvpr}
}

\end{document}